\documentclass[twocolumn]{IEEEtran}
\usepackage{graphicx}

\def\be{\begin{eqnarray}}
\def\en{\end{eqnarray}}
\def\bee{\begin{eqnarray*}}
\def\enn{\end{eqnarray*}}

\newtheorem{remark}{Remark}[section]

\begin{document}

\title{An Analytical Piecewise Radial Distortion Model for Precision Camera Calibration
\thanks{June 20, 2003. For submission to {\em IEEE Trans. on Automation Science and Engineering} as a \underline{\bf Short Paper}. All correspondence should be addressed to Dr. YangQuan Chen. Tel.: 1(435)797-0148, Fax: 1(435)797-3054,  Email: \texttt{yqchen@ece.usu.edu}. CSOIS URL: \texttt{http://www.csois.usu.edu/}} 
}
\author{Lili Ma, {\it Student Member, IEEE}, YangQuan Chen and Kevin L. Moore, {\it Senior Members, IEEE}\\ Center for Self-Organizing and Intelligent Systems (CSOIS),\\Dept. of Electrical and Computer Engineering, 4160 Old Main Hill,\\ Utah State University (USU), Logan, UT 84322-4160, USA.
%\\
%Emails: \texttt{lilima@cc.usu.edu, \{yqchen, moorek\}@ece.usu.edu}
}
\maketitle{}

\begin{abstract}
The common approach to radial distortion is by the means of polynomial approximation, which introduces distortion-specific parameters into the camera model and requires estimation of these distortion parameters. The task of estimating radial distortion is to find a radial distortion model that allows easy undistortion as well as satisfactory accuracy. This paper presents a new piecewise radial distortion model with easy analytical undistortion formula. The motivation for seeking a piecewise radial distortion model is that, when a camera is resulted in a low quality during manufacturing, the nonlinear radial distortion can be complex. Using low order polynomials to approximate the radial distortion might not be precise enough. On the other hand, higher order polynomials suffer from the inverse problem. With the new piecewise radial distortion function, more flexibility is obtained and the radial undistortion can be performed analytically. Experimental results are presented to show that with this new piecewise radial distortion model, better performance is achieved than that using the single function. Furthermore, a comparable performance with the conventional polynomial model using 2 coefficients can also be accomplished. 
\\
\noindent {\bf Key Words:} Camera calibration, Radial distortion, Radial undistortion, Piecewise Function.
\end{abstract}

%%%%%%%%%%%%%%%%%%%%%%%%%%%%%%%%%%%%%%%%%%%%%%%%%%%%%%%%%%%%%%%%%%%%%%%%%%%%%%%%%%%%%%%%%%%%%%%%%%%%%%%%%%%%%
%\thispagestyle{empty}
\section{Introduction}
%%%%%%%%%%%%%%%%%%%%%%%%%%%%%%%%%%%%%%%%%%%%%%%%%%%%%%%%%%%%%%%%%%%%%%%%%%%%%%%%%%%%%%%%%%%%%%%%%%%%%%%%%%%%
Cameras are widely used in many engineering automation processes from visual monitoring, visual metrology to real time visual servoing or visual following. 
%This paper indicates that in precision visual applications, the camera distortion calibration might not be sufficiently precise. 
We will focus on a new camera distortion model which uses a piecewise radial distortion function yet having an analytical undistortion formula, i.e., no numerical iteration is required for undistortion.

\subsection{Camera Calibration}
Camera calibration is to estimate a set of parameters that describes the camera's imaging process. With this set of parameters, a perspective projection matrix can directly link a point in the 3-D world reference frame to its projection (undistorted) on the image plane by:
\begin{equation}
\label{eqn: projection matrix}
\lambda \left [\matrix{u \cr v \cr 1} \right ] 
= {\bf A} \, \left[{\bf R} \mid {\bf t}\right] \left [\matrix{X^w \cr Y^w \cr Z^w \cr 1} \right ] 
= \left[\matrix{\alpha & \gamma &u_0\cr 0 & \beta & v_0\cr 0 & 0 &1}\right] \left [\matrix{X^c \cr Y^c \cr Z^c} \right ],
\end{equation}
where $(u,v)$ is the distortion-free image point on the image plane; the matrix $\bf A$ fully depends on the camera's 5 intrinsic parameters $(\alpha, \gamma, \beta, u_0, v_0)$ with $(\alpha, \beta)$ being two scalars in the two image axes, $(u_0, v_0)$ the coordinates of the principal point, and $\gamma$ describing the skewness of the two image axes; $[X^c, Y^c, Z^c]^T$ denotes a point in the camera frame which is related to the corresponding point $[X^w, Y^w, Z^w]^T$ in the world reference frame by $P^c = {\bf R} P^w + \bf t$ with $({\bf R}, {\bf t})$ being the rotation matrix and the translation vector. For a variety of computer vision applications where camera is used as a sensor in the system, the camera is always assumed fully calibrated beforehand. 

The early works on precise camera calibration, starting in the photogrammetry community, use a 3-D calibration object whose geometry in the 3-D space is required to be known with a very good precision. However, since these approaches require an expensive calibration apparatus, camera calibration is prevented from being carried out broadly. Aiming at the general public, the camera calibration method proposed in \cite{zhang99calibrationinpaper} focuses on desktop vision system and uses 2-D metric information. The key feature of the calibration method in \cite{zhang99calibrationinpaper} is that it only requires the camera to observe a planar pattern at a few (at least 3, if both the intrinsic and the extrinsic parameters are to be estimated uniquely) different orientations without knowing the motion of the camera or the calibration object. Due to the above flexibility, the calibration method in \cite{zhang99calibrationinpaper} is used in this work where the detailed procedures are summarized as: 1) estimation of intrinsic parameters, 2) estimation of extrinsic parameters, 3) estimation of distortion coefficients, and 4) nonlinear optimization.

\subsection{Radial Distortion}
In equation (\ref{eqn: projection matrix}), $(u,v)$ is not the actually observed image point since virtually all imaging devices introduce certain amount of nonlinear distortions. Among the nonlinear distortions, radial distortion, which is performed along the radial direction from the center of distortion, is the most severe part \cite {OlivierF01Straight,tsai87AVersatile}. The radial distortion causes an inward or outward displacement of a given image point from its ideal location. The negative radial displacement of the image points is referred to as the barrel distortion, while the positive radial displacement is referred to as the pincushion distortion \cite{Juyang92distortionmodel}. The removal or alleviation of the radial distortion is commonly performed by first applying a parametric radial distortion model, estimating the distortion coefficients, and then correcting the distortion. 

Lens distortion is very important for accurate 3-D measurement \cite{Tsai88Techniques}.
Let $(u_d, v_d)$ be the actually observed image point and assume that the center of distortion is at the principal point. The relationship between the undistorted and the distorted radial distances is given by 
\begin{equation}
r_d = r + \delta_r,
\end{equation}
where $r_d$ is the distorted radial distance and $\delta_r$ the radial distortion (some other variables used throughout this paper are listed in Table \ref{table: variables used}). 

\begin{table}[htb]
\centering
\caption{List of Variables}
\label{table: variables used}
\renewcommand{\arraystretch}{1.3}
\vspace{-2mm}
{%\small
{\begin {tabular}{|c|l|}\hline
{\bf Variable} & {\bf Description} \\[1ex]\hline
$(u_d, \, v_d)$              & Distorted image point in pixel\\\hline
$(u, \, v)$                  & Distortion-free image point in pixel\\\hline
$(x_d, \, y_d)$              & $[x_d, \, y_d, \, 1]^T = A^{-1} [u_d, \, v_d, \, 1]^T$\\\hline
$(x, \, y)$                  & $[x,\, y,\, 1]^T = A^{-1} [u,\, v,\, 1]^T$ \\\hline
$r_d$                        & $r_d^2 = x_d^2 + y_d^2$ \\\hline
$r$                          & $r^2 = x^2 + y^2$ \\\hline
$\bf k$                      & Radial distortion coefficients \\\hline
\end {tabular}}}   
\end{table}

Most of the existing works on the radial distortion models can be traced back to an early study in photogrammetry \cite{Photogrammetry80} where the radial distortion is governed by the following polynomial equation \cite{zhang99calibrationinpaper,Photogrammetry80,Heikkil97fourstepcameracalibration,Janne96Calibration}:% U-D model in (x,y) 
%\cite{Undistortionchapter,Juyang92distortionmodel}  			    % U-D model in (u,v)
%\cite{GWei94Implicit,zhang96OnThe,Stein97Lens,OlivierF01Straight}    % D-U model in (xd,yd)
%\cite{Janne00Geometric,Prescott97LineBased} 				    % D-U model in (ud,vd)
\begin{equation}
\label{eqn: general polynomial}
r_d = r \, f(r) = r \, (1 + k_1 r^2 + k_2 r^4 + k_3 r^6+ \cdots),
\end{equation}
where $k_1, k_2, k_3, \ldots$ are the distortion coefficients. It follows that 
\begin{equation}
\label{eqn: radial xdyd from xy}
x_d = x \, f(r), \;\;\; y_d = y \, f(r),
\end{equation}
which is equivalent to 
\begin{eqnarray}
\label{eqn: radial (ud,vd) (u,v) relation}
\left\{\hspace{-1mm}
\begin{array}{c}
u_d - u_0 = (u-u_0) \, f(r) \\
v_d - v_0 = (v-v_0) \, f(r)
\end{array}\right.\hspace{-1 mm}.
\end{eqnarray}
This is because
\begin{eqnarray*}
u_d &=& \alpha \, x_d + \gamma \, y_d + u_0, 		\\
    &=& \alpha \, x f(r) + \gamma \, y f(r) + u_0 	\\
    &=& (u-u_0) \, f(r) + u_0 				\\
v_d &=& \beta \, y_d  + v_0.					\\
    &=& (v-v_0) \, f(r) + v_0					
\end{eqnarray*}

For the polynomial radial distortion model (\ref{eqn: general polynomial}) and its variations, the distortion is especially dominated by the first term and it has also been found that too high an order may cause numerical instability \cite{tsai87AVersatile,zhang99calibrationinpaper,GWei94Implicit}. When using two coefficients, the relationship between the distorted and the undistorted radial distances becomes \cite{zhang99calibrationinpaper}
\begin{eqnarray}
\label{eqn: radial distortion order 2 4}
r_d = r \, (1 + k_1 \, r^2 + k_2 \, r^4).
\end{eqnarray}
The inverse of the polynomial function in (\ref{eqn: radial distortion order 2 4}) is difficult to perform analytically but can be obtained numerically via an iterative scheme. In \cite{Undistortionchapter}, for practical purpose, only one distortion coefficient $k_1$ is used. 

To overcome the inversion problem, another polynomial radial distortion model using also two terms is presented in \cite{LiliISIC03Flex} with
\be \label{eqn: polynomial 1 2} f(r) = 1 + k_1 \, r + k_2 \, r^2, \en
whose appealing feature lies in its satisfactory accuracy as well as the existence of an easy analytical undistortion formula. The polynomial radial distortion model in (\ref{eqn: polynomial 1 2}), together with the commonly used model (\ref{eqn: radial distortion order 2 4}), acts as the benchmark for evaluating the performance of the piecewise radial distortion model proposed in Sec.~\ref{sec: piecewise 2nd order}. 

In this work, a new piecewise radial distortion model is proposed, which allows analytical radial undistortion along with preserving high calibration accuracy. To compare the performance of different distortion models, final value of optimization function $J$, which is defined to be \cite{zhang99calibrationinpaper}:
\begin{equation}
\label{eqn: objective function}
J = \sum_{i=1}^N \sum_{j=1}^n \|m_{i j}-\hat m({\bf A}, {\bf k}, {\bf R}_i, {\bf t}_i, M_j) \|^2,
\end{equation}
is used, where $\hat m({\bf A}, {\bf k}, {\bf R}_i, {\bf t}_i, M_j)$ is the projection of point $M_j$ in the $i^{th}$ image using the estimated parameters; $\bf k$ denotes the distortion coefficients; $M_j$ is the $j^{th}$  3-D point in the world frame with $Z^w = 0$; $n$ is the number of feature points in the coplanar calibration object; and $N$ is the number of images taken for calibration.
In \cite{zhang99calibrationinpaper}, the estimation of radial distortion is done after having estimated the intrinsic and the extrinsic parameters and just before the nonlinear optimization step. So, for different radial distortion models, we can reuse the estimated intrinsic and extrinsic parameters. 

The rest of the paper is organized as follows. Sec.~\ref{sec: Polynomial} describes the polynomial radial distortion model (\ref{eqn: polynomial 1 2}) with its analytical undistortion function. In Sec.~\ref{sec: piecewise 2nd order}, the new piecewise radial distortion model is proposed. Comparisons with models (\ref{eqn: radial distortion order 2 4}) and (\ref{eqn: polynomial 1 2}) are presented in Sec.~\ref{sec: experimental} in detail. Some concluding remarks are given in Sec.~\ref{sec: conclusion}. Finally, a dedicated section, Sec.~\ref{notes_to_practitioners}, is prepared for practitioners on how this work can be useful in practice.

%%%%%%%%%%%%%%%%%%%%%%%%%%%%%%%%%%%%%%%%%%%%%%%%%%%%%%%%%%%%%
\section{Radial Distortion Model (\ref{eqn: polynomial 1 2})}
\label{sec: Polynomial}
%%%%%%%%%%%%%%%%%%%%%%%%%%%%%%%%%%%%%%%%%%%%%%%%%%%%%%%%%%%%%

\subsection{Model}

The conventional radial distortion model (\ref{eqn: radial distortion order 2 4}) with 2 parameters does not have an exact inverse, though there are ways to approximate it without iterations, such as the model described in \cite{Janne00Geometric}, where $r$ can be calculated from $r_d$ by
\begin{equation}
\label{eqn: undistortion approximation}
r =  r_d \, ( 1 - k_1 \, r_d^2 - k_2 \, r_d^4).
\end{equation}
The fitting results given by the above model can be satisfactory when the distortion coefficients are small values. However,   equation (\ref{eqn: undistortion approximation}) introduces another source of error that will inevitably degrade the calibration accuracy. Due to this reason, an analytical inverse function that has the advantage of giving the exact undistortion solution is one of the main focus of this work.

To overcome the shortcoming of no analytical inverse formula but still preserving a comparable accuracy, the radial distortion model (\ref{eqn: polynomial 1 2}) has the following three properties:
\begin{itemize}
\item[{\rm \bf 1)}] This function is radially symmetric around the center of distortion (which is assumed to be at the principal point $(u_0, v_0)$ for our discussion) and it is expressible in terms of radius $r$ only;
\item[{\rm \bf 2)}] This function is continuous, hence $r_d = 0$ iff $r = 0$;
\item[{\rm \bf 3)}] The resultant approximation of $x_d$ is an odd function of $x$, as can be seen next.
\end{itemize}
Introducing a quadratic term $k_1 \, r^2$ in (\ref{eqn: polynomial 1 2}), this distortion model still approximates the radial distortion, since the distortion is in the radial sense. 

From (\ref{eqn: polynomial 1 2}), we have
\begin{equation}
\left \{\hspace{-1mm}
\begin{array}{l}
x_d = x \, f(r) = x \, (1 + k_1 r + k_2 r^2)  \\
y_d = y \, f(r) = y \, (1 + k_1 r + k_2 r^2)
\end{array}\right.\hspace{-1 mm}.
\end{equation}
It is obvious that $x_d = 0$ iff $x = 0$. When $x_d \ne 0$, by letting $c = y_d/x_d = y /x$, we have $y = cx$ where $c$ is a constant. Substituting $y = cx$ into the above equation gives
\begin{eqnarray}
\label{eqn: distortion model 3}
x_d &=& x \, \left[ 1+k_1 \sqrt{x^2 + c^2x^2} + k_2(x^2 + c^2x^2)\right] \nonumber\\
   &=& x \, \left[ 1+k_1 \sqrt{1 + c^2} \, {\tt sgn}(x) x + k_2(1 + c^2)x^2 \right] \nonumber\\
   &=& x + k_1 \sqrt{1 + c^2} \, {\tt sgn}(x) \, x^2 + k_2(1 + c^2) \, x^3,
\end{eqnarray}
where ${\tt sgn}(x)$ gives the sign of $x$ and $x_d$ is an odd function of $x$. 

The well-known radial distortion model (\ref{eqn: general polynomial}) that describes the laws governing the radial distortion does not involve a quadratic term. 
%Besides, model (\ref{eqn: general polynomial}) has solid physical basis and has been verified with precise ray tracing.
Thus, it might be unexpected to add one. However, when interpreting from the relationship between $(x_d,y_d)$ and $(x,y)$ in the camera frame as in equation (\ref{eqn: distortion model 3}), the radial distortion function is to approximate the $x_d \leftrightarrow x$ relationship which is intuitively an odd function. Adding a quadratic term to $\delta_r$ does not alter this fact. Furthermore, introducing quadratic terms to $\delta_r$ broadens the choice of radial distortion functions. 

\begin{remark}
The radial distortion models discussed in this paper belong to the category of {\underline U}ndistorted-{\underline D}istorted model, while the {\underline D}istorted-{\underline U}ndistorted model also exists in the literature to correct the distortion \cite{Toru02Unified}. The radial distortion models can be applied to the D-U formulation simply by defining
\begin{displaymath}
r = r_d \, f(r_d).
\end{displaymath}
Consistent results and improvement can be achieved in the above D-U formulation. 
\end{remark}

%%%%%%%%%%%%%%%%%%%%%%%%%%%%%%%%%%%%%%%%%%%%%%%%%
\subsection{Radial Undistortion of Model (\ref{eqn: polynomial 1 2})}
\label{sec: undistortion}

From (\ref{eqn: polynomial 1 2}), we have
\begin{displaymath}
r^3 + a \, r^2 + b \, r + c = 0,
\end{displaymath}
with $a = k_1/k_2, b = 1/k_2$, and $c = -r_d/k_2$. Let ${\bar r} = r - a/3$, the above equation becomes
\begin{displaymath}
{\bar r}^3 + p \, {\bar r} + q = 0,
\end{displaymath}
where $p = b - a^2/3, \; q = 2a^3/27 - a b/3 + c$. Let $\Delta = (\frac{q}{2})^2 + (\frac{p}{3})^3$. If $\Delta > 0$, there is only one solution; if $\Delta = 0$, then $r = 0$ , which occurs when $\delta_r = 0$; if $\Delta < 0$, then there are three solutions. In general, the middle one is what we need, since the first root is at a negative radius and the third lies beyond the positive turning point \cite{zhang96OnThe,Ben02PhD}. % Page_2 of zhang96 and Page_24 of the Ph.D thesis
After $r$ is determined, $(u,v)$ can be calculated from (\ref{eqn: radial (ud,vd) (u,v) relation}) uniquely.

%%%%%%%%%%%%%%%%%%%%%%%%%%%%%%%%%%%%%%%%%%%%%%%%%%%%%
\section{Piecewise Radial Distortion Model}
\label{sec: piecewise 2nd order}

\begin{figure}[tb]
\centering
\includegraphics[width=0.4\textwidth]{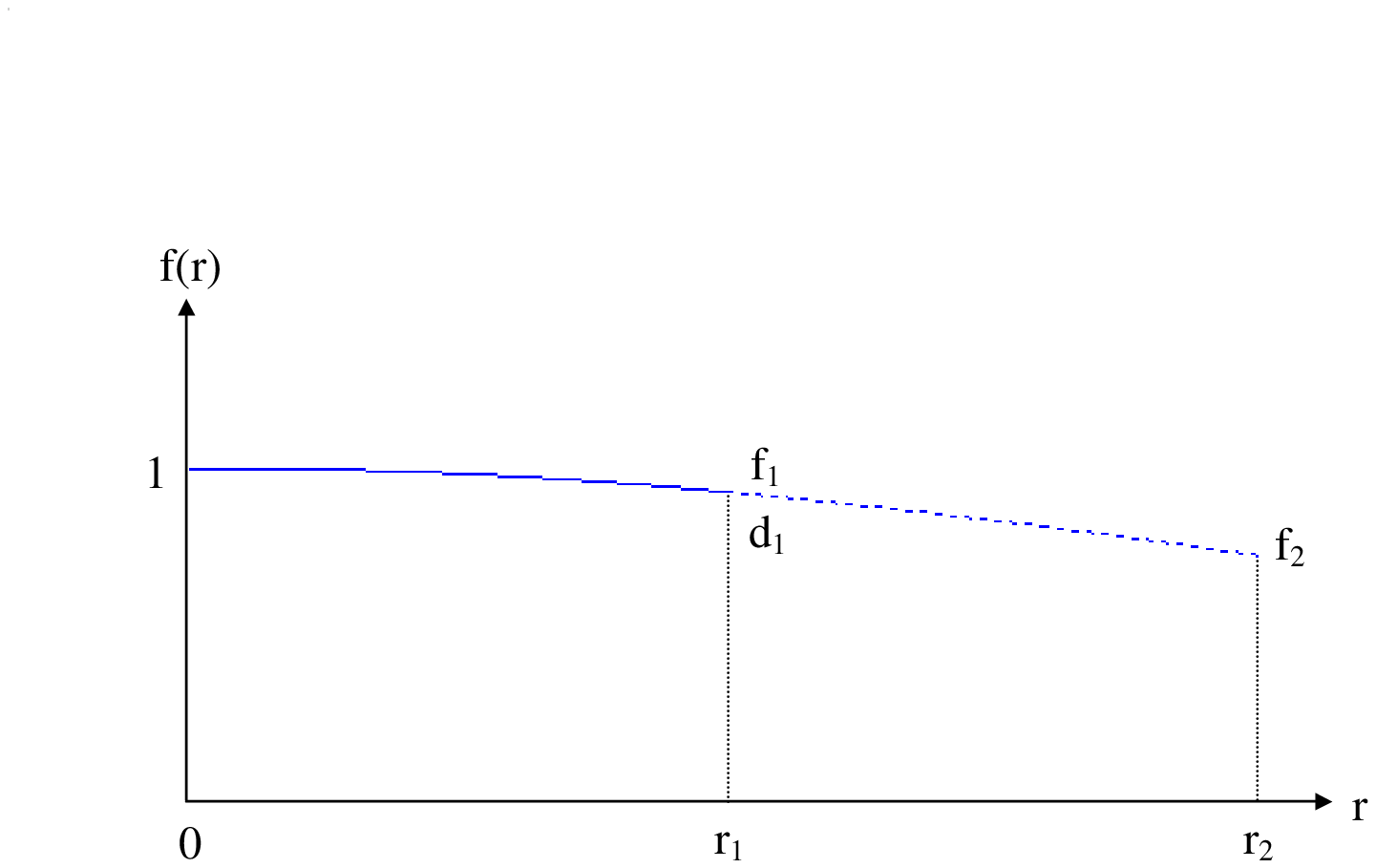}
\caption {A smooth piecewise function (two-segment).}
\label{fig: illustration}
\end{figure}

A two-segment radial distortion function is proposed and illustrated in Fig.~\ref{fig: illustration}, where each segment is a function of the form
\begin{equation} \label{eqn: piecewise model}
\left \{\hspace{-1mm}
\begin{array}{l}
f_1(r) = a_0 + a_1 \, r + a_2 \, r^2, \;\; {\rm for} \;\; r \in [0, r_1]\\[4pt]
f_2(r) = b_0 + b_1 \, r + b_2 \, r^2, \;\;\; {\rm for} \;\; r \in (r_1, r_2]
\end{array}\right.\hspace{-1 mm},
\end{equation}
with $r_1 = r_2/2$. We are interested in estimating the coefficients $(a_0, a_1, a_2)$ and $(b_0, b_1, b_2)$ such that the two polynomials are continuous and smooth at the interior knot $r = r_1$. The reason for choosing a distortion function in (\ref{eqn: polynomial 1 2}) for each segment is that the radial undistortion can be performed using the analytical procedures described in Sec.~\ref{sec: undistortion} with no iterations. 

To ensure that the overall function (\ref{eqn: piecewise model}) is continuous and smooth across the interior knot, the following 6 constraints can be applied
\begin{eqnarray}
\label{eqn: 6 constraints}
\left\{
\begin{array}{r}
f_1(0) = 1 \;\,                        \\[2pt]
a_0 + a_1 \, r_1 + a_2 \, r_1^2 = f_1  \\[2pt]
a_1 + 2 a_2 \, r_1 = d_1               \\[2pt]
b_0 + b_1 \, r_1 + b_2 \, r_1^2 = f_1  \\[2pt]
b_1 + 2 b_2 \, r_1 = d_1               \\[2pt]
b_0 + b_1 \, r_2 + b_2 \, r_2^2 = f_2
\end{array}\right.\hspace{-1mm},
\end{eqnarray}
where $f_1 = f_1(r_1) = f_2(r_1)$, $f_2 = f_2(r_2)$, and $d_1 = {\dot f}_1(r_1) = {\dot f}_2(r_1)$. By enforcing that the two segments have the same value and derivative at the interior knot $r_1$, the resultant single function is guaranteed to be continuous and smooth over the whole range $[0, r_2]$. Since each interior knot provides 4 constraints to make the resultant single function smooth, in order to estimate the coefficients $(a_0, a_1, a_2)$ and $(b_0, b_1, b_2)$ uniquely, we need another two constraints, which are chosen to be $f_1(0)$ and $f_2(r_2)$ in (\ref{eqn: 6 constraints}).

Since the coefficients $(a_0, a_1, a_2)$ and $(b_0, b_1, b_2)$ in (\ref{eqn: 6 constraints}) can be calculated uniquely from $(f_1, d_1, f_2)$ by
\begin{equation} \label{eqn: a b solutions}
\left\{
\begin{array}{l}
a_0 = 1\\[2pt]
a_1 = (-2 - r_1 d_1 + 2 f_1) \,/\, r_1\\[2pt]
a_2 =  (1 + r_1 d_1 -   f_1) \,/\, r_1^2\\[2pt]
b_2 = (f_2 - f_1 + r_1 d_1 - r_2 d_1)\,/\,(r_1 - r_2)^2\\[2pt]
b_1 = d_1 - 2 \,b_2 r_1\\[2pt]
b_0 = f_1 - d_1 r_1 + b_2 r_1^2
\end{array}
\right.\hspace{-1mm},
\end{equation}
the radial distortion coefficients that are used in the nonlinear optimization for the piecewise radial distortion model can be chosen to be $(f_1, d_1, f_2)$ with the initial values $(1, 0, 1)$, which has only one extra coefficient compared with the single model (\ref{eqn: polynomial 1 2}). During the nonlinear optimization process, the coefficients $(a_0, a_1, a_2)$ and $(b_0, b_1, b_2)$ are calculated from (\ref{eqn: a b solutions}) in each iteration. 

The purpose of this work is to show that the proposed piecewise radial distortion model achieves the following properties: 
\begin{itemize}
\item [\bf 1)] Given $r_d$ and the distortion coefficients, the solution of $r$ from $r_d$ has closed-form solution; 
\item [\bf 2)] It approximates the commonly used distortion model (\ref{eqn: radial distortion order 2 4}) with higher accuracy than the single function (\ref{eqn: polynomial 1 2}) based on the final value of $J$ in (\ref{eqn: objective function}).
\end{itemize}

%%%%%%%%%%%%%%%%%%%%%%%%%%%%%%%%%%%%%%%%%%%%%%
\section{Experimental Results and Discussions}
\label{sec: experimental}

\begin{figure*}[htb]
\centering
\includegraphics[width=1.0\textwidth]{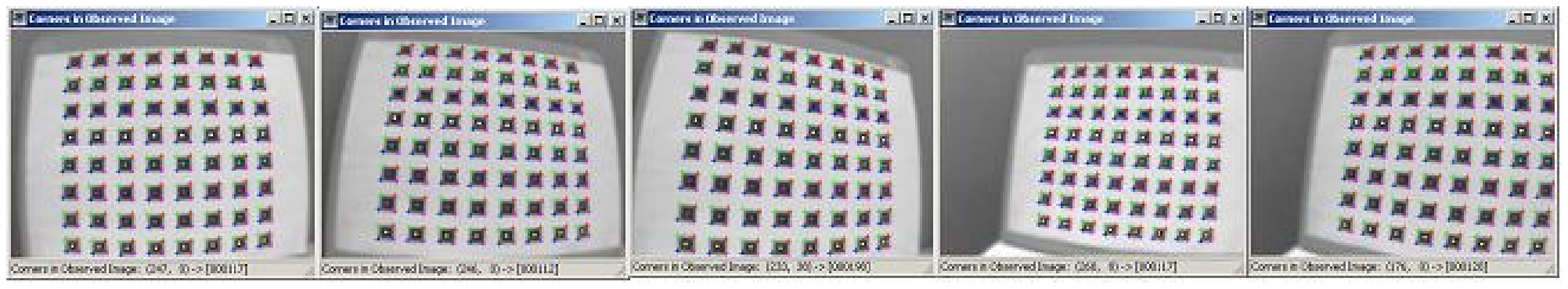}
\caption {Five images of the model plane with the extracted corners (indicated by cross) for the desktop camera.}
\label{fig: extracted desktop}
\end{figure*}

\begin{figure*}[htb]
\centering
\includegraphics[width=1.0\textwidth]{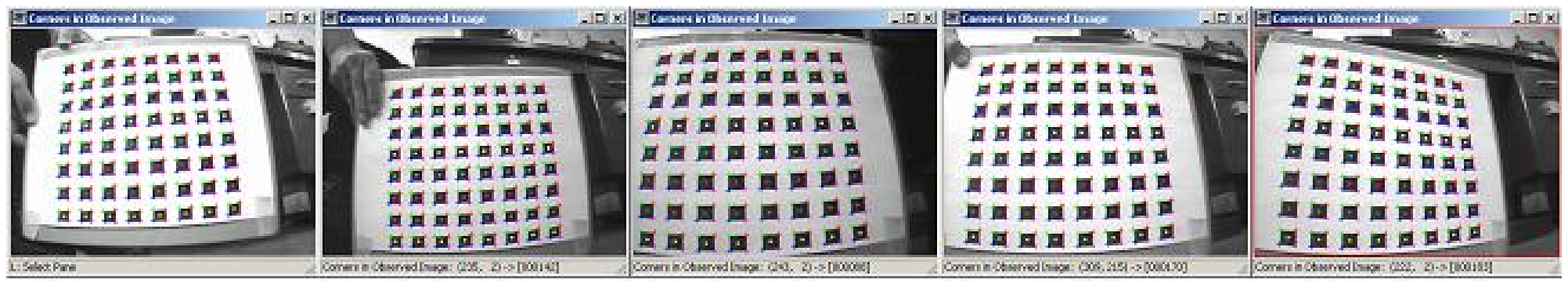}
\caption {Five images of the model plane with the extracted corners (indicated by cross) for the ODIS camera.}
\label{fig: extracted ODIS}
\end{figure*}

For the two-segment piecewise distortion model (\ref{eqn: piecewise model}), comparisons are made with the single model (\ref{eqn: polynomial 1 2}) and the commonly used model (\ref{eqn: radial distortion order 2 4}) based on the final value of the objective function $J$ in (\ref{eqn: objective function}) after nonlinear optimization by the Matlab function {\tt fminunc}, since common approach to camera calibration is to perform a full-scale nonlinear optimization for all parameters.
Using the public domain testing images \cite{zhang98calibrationwebpage}, the desktop camera images \cite{Lilicalreport02} (a color camera in our CSOIS), and the ODIS camera images \cite{Lilicalreport02,odiscamera} (the camera on ODIS robot built in our CSOIS), the final values of $J$, the estimated distortion coefficients, and the 5 estimated intrinsic parameters ($\alpha, \beta, \gamma, u_0, v_0$), are summarized in Table~\ref{table: comparison results}, where the listed distortion coefficients are $(k_1, k_2)$ for the single models (\ref{eqn: radial distortion order 2 4}), (\ref{eqn: polynomial 1 2}) and $(f_1, d_1, f_2)$ for the piecewise. The extracted corners for the model plane of the desktop and the ODIS cameras are shown in Figs.~\ref{fig: extracted desktop} and \ref{fig: extracted ODIS}. As noticed from these images, the two cameras both experience a barrel distortion. The plotted dots in the center of each square are only used for judging the correspondence with the world reference points. 

From Table~\ref{table: comparison results}, it is observed that the values of $J$ using the piecewise model for the three groups of testing images are always less than those using the single function (\ref{eqn: polynomial 1 2}). Furthermore, the fitting residuals are closer to those of model (\ref{eqn: radial distortion order 2 4}). 
For the ODIS images, the value of $J$ of the piecewise model is even smaller than that of model (\ref{eqn: radial distortion order 2 4}). 
Thus, it is concluded that improvement has been achieved using the proposed piecewise radial distortion model. 
The comparison between  model (\ref{eqn: piecewise model}) with models (\ref{eqn: radial distortion order 2 4}), (\ref{eqn: polynomial 1 2}) might not be fair since the new model has one more coefficient.
%and it is evident that each additional coefficient in the model tends to decrease the fitting residual. 
However, our main point is to emphasize that by applying the piecewise idea, higher accuracy can be achieved without sacrificing the property of having analytical undistortion function. 

\begin{remark}
Classical criteria that are used in the computer vision to assess the accuracy of calibration includes the radial distortion as one part inherently \cite{Juyang92distortionmodel}. However, to our best knowledge, there is not a systematically quantitative and universally accepted criterion in the literature for performance comparisons among different radial distortion models. Due to this lack of criterion, in our work, the comparison is based on, but not restricted to, the fitting residual of the full-scale nonlinear optimization in (\ref{eqn: objective function}). 
\end{remark}

\begin{table*}[htb]
\centering
\caption{Comparison of Radial Distortion Models Using Three Groups of Testing Images}
\label{table: comparison results}
\renewcommand{\arraystretch}{1}
\setlength{\tabcolsep}{1.8mm}
\vspace{-2mm}
{%\small
{\begin {tabular}{|c|c|c|rrr|rrrrr|}\hline
{\bf Images} &{\bf Model}& {$J$} & \multicolumn{3}{|c|}{\bf Distortion Coefs} & \multicolumn{5}{|c|}{\bf Intrinsic Parameters $(\alpha, \gamma, u_0, \beta, v_0)$} \\\hline
&(\ref{eqn: radial distortion order 2 4})&144.8802&-0.2286&0.1905&-&832.4860&0.2042&303.9605&832.5157&206.5811\\\cline{2-11}
\bf Public&(\ref{eqn: polynomial 1 2})&145.6592&-0.0215&-0.1566&-&833.6508& 0.2075&303.9847&833.6866&206.5553\\\cline{2-11}
&(\ref{eqn: piecewise model})&144.8874&0.9908&-0.0936&0.9653&831.7068&0.2047& 303.9738 & 831.7362 & 206.5670\\\hline\hline
&(\ref{eqn: radial distortion order 2 4})&778.9767&-0.3435&0.1232&-&277.1449&-0.5731&153.9882&270.5582&119.8105\\\cline{2-11}
\bf Desktop&(\ref{eqn: polynomial 1 2})&803.3074&-0.1067&-0.1577&-&282.5642&-0.6199&154.4913&275.9019&120.0924\\\cline{2-11}
&(\ref{eqn: piecewise model})&782.5865&0.9387&-0.2695&0.8066&277.4852&-0.5757&154.0058&270.9052 & 119.7416\\\hline\hline
&(\ref{eqn: radial distortion order 2 4})&840.2650&-0.3554&0.1633&-&260.7658&-0.2741&140.0581&   255.1489&113.1727\\\cline{2-11}
\bf ODIS&(\ref{eqn: polynomial 1 2})&851.2619&-0.1192&-0.1365& -&266.0850&-0.3677&139.9198&260.3133&113.2412\\\cline{2-11}
&(\ref{eqn: piecewise model})&838.5678&0.9410&-0.2563&0.8270&261.9485&-0.2875 & 140.2521 & 256.3134 & 113.0856\\\hline
\end {tabular}}}
\end{table*}

\begin{figure}[htb]
\centering
\includegraphics[width=0.37\textwidth]{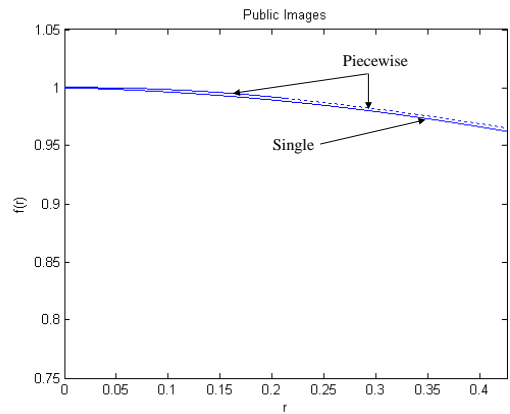}
\caption {$f(r)$ curves for the public images.}
\label{fig: f(r) m}
\end{figure}

\begin{figure}[htb]
\centering
\includegraphics[width=0.37\textwidth]{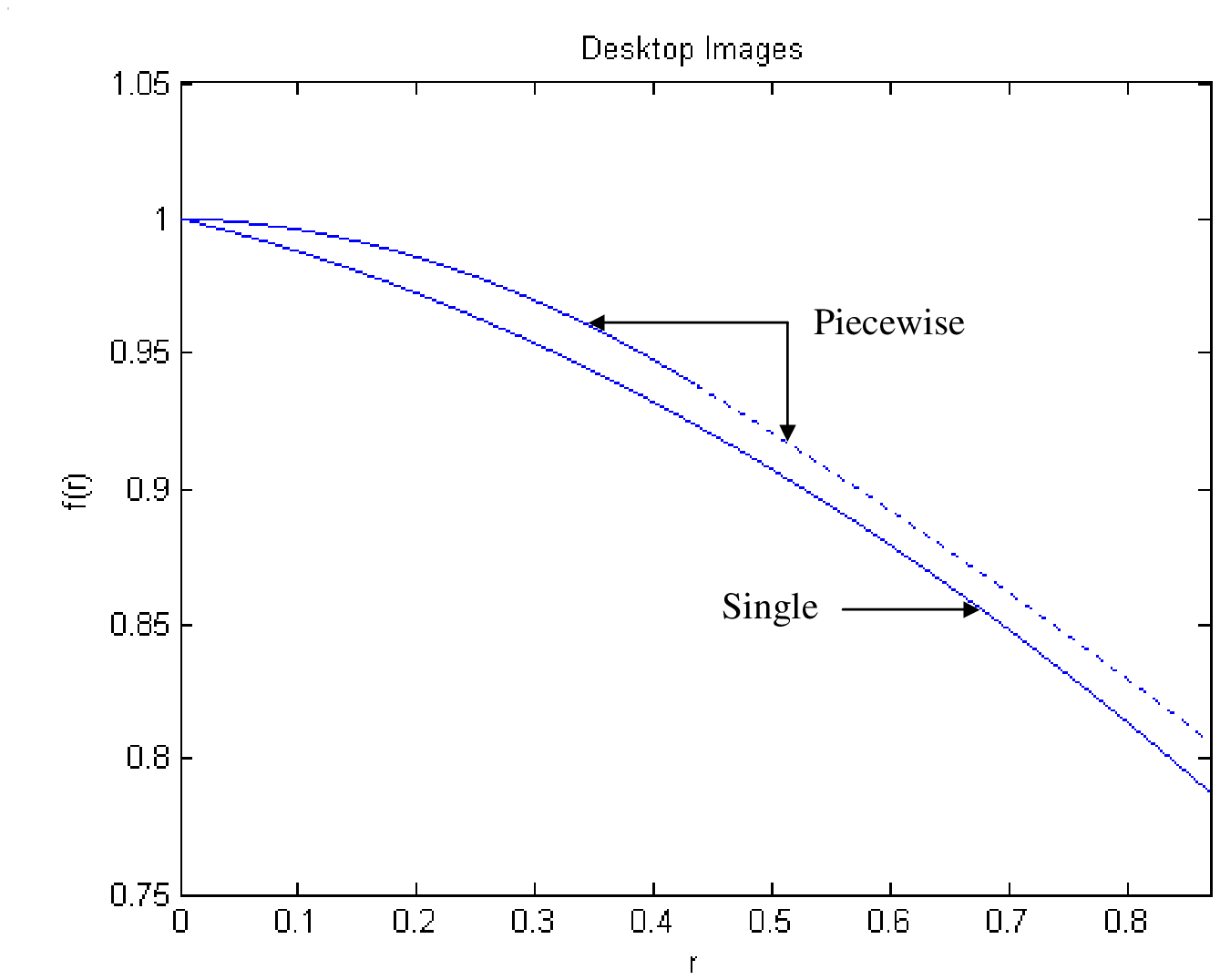}
\caption {$f(r)$ curves for the desktop images.}
\label{fig: f(r) d}
\end{figure}

\begin{figure}[htb]
\centering
\includegraphics[width=0.37\textwidth]{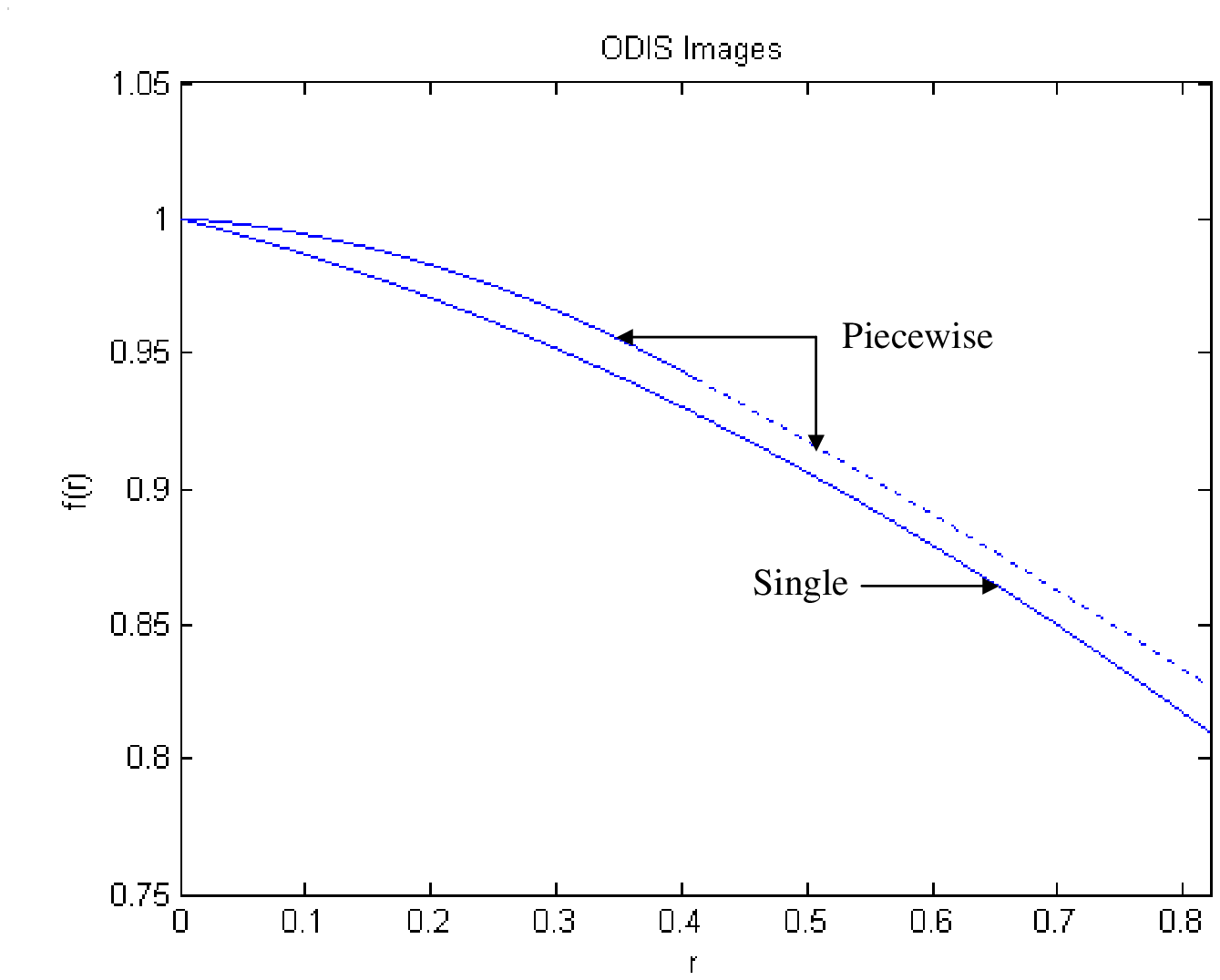}
\caption {$f(r)$ curves for the ODIS images.}
\label{fig: f(r) o}
\end{figure}

The resultant estimated $f(r)$ curves for the three groups of testing images using the piecewise model (\ref{eqn: piecewise model}) are plotted in Figs.~\ref{fig: f(r) m}, \ref{fig: f(r) d}, and \ref{fig: f(r) o}, respectively, where the $f(r)$ curves using the single function (\ref{eqn: polynomial 1 2}) are also plotted for comparison. It is observed that the three piecewise $f(r)$ curves are all slightly above the single $f(r)$ curves. Unfortunately, so far, it is not clear to us whether or not it happens to be this case. 
%It seems that the radial distortion is exaggerated when using inappropriate approximation models. From our work, we have the following observations:
%\begin{itemize}
%\item [\rm \bf 1)] Compared with the single quadratic radial distortion model (\ref{eqn: polynomial 1 2}), using the %proposed piecewise quadratic model (\ref{eqn: piecewise model}) the resultant estimated %parameters, including the intrinsic parameters, the extrinsic parameters, and the distortion coefficients, are closer to the true values, since the value of $J$ is decreased;
%\item [\rm \bf 2)] Using the common approach to perform a full-scale optimization for all parameters by minimizing $J$, the %selection of the radial distortion model not only affects whether or not the distortion can be estimated properly, but also %affects whether or not the other camera parameters can be estimated more accurately.
%\end{itemize}

One issue in the implementation of the nonlinear optimization is how to decide $r_2$, which is related to the estimated extrinsic parameters that are changing from iteration to iteration during the nonlinear optimization process. In our implementation, for each camera, 5 images are taken. $r_2$ is chosen to be the maximum $r$ of all the extracted feature points on the 5 images for each iteration.

\begin{remark}
To make the results in this paper reproducible by other researchers for further investigation, we present the options we use for the nonlinear optimization: \texttt{options = optimset(`Display', `iter', `LargeScale', `off', `MaxFunEvals', 8000, `TolX', $10^{-5}$,  `TolFun', $10^{-5}$, `MaxIter', 120)}. The raw data of the extracted feature locations in the image plane are also available \cite{Lilicalreport02}. 
\end{remark}

%%%%%%%%%%%%%%%%%%%%%%%%%%%%
\section{Concluding Remarks}   
\label{sec: conclusion}

This paper proposes a new piecewise polynomial radial distortion model for camera calibration. The appealing part of this piecewise model is that it preserves high accuracy and the property of having analytical undistortion formula for each segment. Experiments results are presented to show that this new piecewise radial distortion model can be quite accurate and performance improvement is achieved compared with the corresponding single radial distortion function. Furthermore, a comparable performance with the conventional polynomial radial distortion model using 2 coefficients can also be accomplished. 

%%%%%%%%%%%%%%%%%%%%%%%%%%%%%%%
\section{Note to Practitioners}
\label{notes_to_practitioners}

In precision automation applications, cameras   are widely used. However, from the results reported in this paper, the camera distortion  calibration is usually not well done as demonstrated in our ODIS (Omnidirectional Inspection System, \texttt{http://www.csois.usu.edu}) camera. We thought that the ODIS camera has a good quality but in fact it is not at least for the camera distortion property. To ensure the precision of camera calibration along with the property of having analytical inverse function, the distortion  model should be re-visited by using the method of this paper. For noncommercial use, all relevant code of this work is available from \cite{Lilicalreport02}. 

%%%%%%%%%%%%%%%%%%%%%%%%%%%%%%%%%%%%%%%%
\bibliography{D:/work/cameracalibration/calibration,csois1,csois2}
\end{document}